# 3D Pose Estimation of Tomato Peduncle Nodes using Deep Keypoint Detection and Point Cloud

Jianchao Ci[a,*], Xin Wang[a], David Rapado-Rincón[a], Akshay K. Burusa[a], Gert Kootstra[a]

[a]*Agricultural Biosystems Engineering Group, Department of Plant Sciences, Wageningen University and Research, P.O. Box 16, Wageningen, 6700AA, the Netherlands*

## Abstract

Greenhouse production of fruits and vegetables in developed countries is challenged by labor scarcity and high labor costs. Robots offer a good solution for sustainable and cost-effective production. Acquiring accurate spatial information about relevant plant parts is vital for successful robot operation. Robot perception in greenhouses is challenging due to variations in plant appearance, viewpoints, and illumination. This paper proposes a keypoint-detection-based method using data from an RGB-D camera to estimate the 3D pose of peduncle nodes, which provides essential information to harvest the tomato bunches.

Specifically, this paper proposes a method that detects four anatomical landmarks in the color image and then integrates 3D point-cloud information to determine the 3D pose. A comprehensive evaluation was conducted in a commercial greenhouse to gain insight into the performance of different parts of the method. The results showed: (1) high accuracy in object detection, achieving an Average Precision (AP) of AP@0.5=0.96; (2) an average Percentage of Detected Joints (PDJ) of the keypoints of PhDJ@0.2=94.31%; and (3) 3D pose estimation accuracy with mean absolute errors (MAE) of $11.38°$ and $9.93°$ for the relative upper and lower angles between the peduncle and main stem, respectively. Furthermore, the capability to handle variations in viewpoint was investigated, demonstrating the method was robust to view changes. However, canonical and higher views resulted in slightly higher performance compared to other views. Although tomato was selected as a use case, the proposed method is also applicable to other greenhouse crops like pepper.

# 1. Introduction

Growing vegetables in greenhouses can significantly extend the production period of plants, which increases yields and brings economic benefits to the owner of the greenhouse. Most greenhouse crops, such as tomatoes, cucumber, and bell pepper require selective maintenance and harvesting, making production activities labor-intensive. This poses a huge challenge for production because the labor cost is high and labor is scarce (Benavides et al., 2020). A higher level of automation and robotization is seen as a good solution to replace human labor in greenhouse production, and many studies have been proposed to more effectively involve robotics in greenhouse operations, such as harvesting (Bac et al., 2014; Ji et al., 2012; Yoshida et al., 2018), monitoring (Halstead et al., 2018), and phenotyping (Boogaard et al., 2020; Virlet et al., 2016; Vit et al., 2020).

Most greenhouse operations require the robotic perception system to detect the relevant parts of the plant (e.g., fruits, peduncles, and main stem) and to collect sufficient information about them to perform manipulations, such as harvesting, deleafing, and pruning. The perception process is adversely affected by variation in the appearance of the plant and occlusions that are significantly present in complex greenhouse environments (Afonso et al., 2019; Kootstra et al., 2021). Variation results from natural variation in the growth of plants, creating differences in the morphology and appearance of the plant parts, as well as from environmental influences, such as changing illumination. Occlusions happen frequently in the cluttered greenhouse environments, where plant parts are frequently entirely or partially obstructed by other plant parts.

To overcome the challenge of variation and improve perception performance, in recent years, deep-learning-based vision systems have been extensively used in agricultural scenarios. Compared to traditional methods that use handcrafted features, deep-learning methods include the feature extraction as part of the end-to-end learning process, which has been shown to perform much better in terms of accuracy and robustness to variation. Kamilaris and Prenafeta-Boldú (2018) reviewed 40 studies using deep learning in agricultural applications and concluded that deep neural networks are superior to other methods. Bargoti and Underwood (2017), for instance, used a deep neural network for object detection of several different fruits, including mangoes, apples, and almonds in orchards, while Sa et al. (2016) used a similar method on a combination of color and near infra-red images to successfully detect seven different fruit types in greenhouse environments. Boogaard et al. (2020) used deep object detection to detect the leaf and fruit nodes of cucumber plants. A deep neural network for instance segmentation of grape bunches in orchards was used in (Santos et al., 2020). The same method was used in (Shi et al., 2019) to segment images of tomato seedlings in stem, node, and the individual leaves. Kang and Chen (2020) proposed a new network for instance segmentation of apples. These detection algorithms are able to locate different plant parts in 2D images. However, to perform a robotic operation, 3D information about the position and orientation of the objects is needed so that the robot can bring a tool to the desired location and orientation.

Often, the 3D pose of an object is defined by the 3D position and the 3D orientation of the object. This was used, for instance in (Eizentals & Oka, 2016; Lehnert et al., 2016) to represent the pose of bell peppers. In both studies, the pose of the fruit was estimated by fitting a 3D template to the acquired partial point cloud of the fruit. Li et al. (2018) estimated the pose of peppers by finding the symmetry axes. A deep neural network was proposed in Wagner et al. (2022) to estimate the 3D pose of strawberries directly from the color-and-depth (RGB-D) image. Kang et al. (2020) proposed a point-based neural network to learn to estimate the grasp pose directly from the point cloud of the fruit.

In the animal domain, a different definition of object pose is used. There, computer-vision methods, inspired by human-pose detectors, estimate the pose of the animal by a set of keypoints, representing anatomical landmarks on the animal (e.g., (X. Li et al., 2019; Mathis et al., 2020; Pereira et al., 2019; Russello et al., 2022). The pose provides information on the location of these keypoints, as well as the relations between the keypoints. The advantages of this representation are that it can be used for articulated objects and that they are more robust to occlusion, as the non-occluded keypoints can still be detected to represent the object pose. This makes it interesting for pose estimation of plant parts but it has hardly been explored to date, with the exception of Zhang et al. (2022) who used keypoint detection to get the pose of tomato bunches using a combination of stacked hourglasses and an object-detection network. The method localized a set of 2D keypoints on the tomatoes, peduncle and stem and then converting them into 3D keypoints using spatial information provided by the depth camera. The accuracy of the 3D-pose estimation, however, was not quantitatively evaluated, so it is unclear if the method can be used for robotic operation.

This study focusses on the 3D pose estimation of peduncle nodes on tomato plants to provide information for fruit harvesting. The peduncle connects the fruit with the main stem. To avoid fruit damage and extend shelf life, tomato trusses are normally detached from the plant by cutting the peduncle. Some other studies tried to detect and localize the peduncle relying on the detection of other plant parts (such as the fruits and branches), then determining the peduncle according to specific morphological relationships between them (Sa et al., 2017; Yoshida et al., 2020). The downside of these methods is that they pose assumptions on the pose of the peduncle. Some applications, for instance, assumed the peduncle to be on top of the fruit, requiring the fruit cluster to be downwards and vertical (Liang et al., 2020; Luo et al., 2016). This limits the application, as the tomato peduncles can be oriented in any direction. In (Boogaard et al., 2020), nodes on cucumber plants were directly detected in the images by a deep neural network (DNN) without assumptions on the relations to other plant parts, however, there work only detected the position of the nodes, not the full pose needed for robotic operations. In (Rong et al., 2022), a DNN for instance-segmentation method was used to predict an image mask for tomato peduncles, after which a handcrafted algorithm was used to determine the 3D cutting position. Kim et al. (2023) successfully utilized an end-to-end approach to detect keypoint on images taken in a tomato greenhouse, but their work was limited by predicting only the 2D pose of the pedicels connecting single tomatoes to the truss. In our work, we combined the benefits of an end-to-end pose detection network with the benefits of including 3D information to estimate the full 3D pose of peduncles, enabling robotic operations.

In this paper, we proposed a keypoint-detection-based method to estimate the 3D pose of the peduncle nodes of tomato plants based on data from an RGB-D camera. The 3D pose was defined by four points: the node, a point on the peduncle and two points up and down the main stem. As shown in Figure 1, the method was built upon three steps: (1) a deep keypoint detector was used for robust detection of the peduncle node and its pose in 2D images, (2) the detected keypoints were projected to the aligned 3D point cloud, and (3) the 3D pose was estimated. The method was thoroughly evaluated on three aspects: the node detection, the keypoint detection, and the accuracy of the 3D pose compared to ground-truth measurements. The first two analyses were done on the image level, while the last analysis evaluated the 3D pose estimation in the world coordinate system compared to manual measurements. Because the success of peduncle node detection might depend on the viewpoint, we analyze results for different camera poses with respect to the target object.

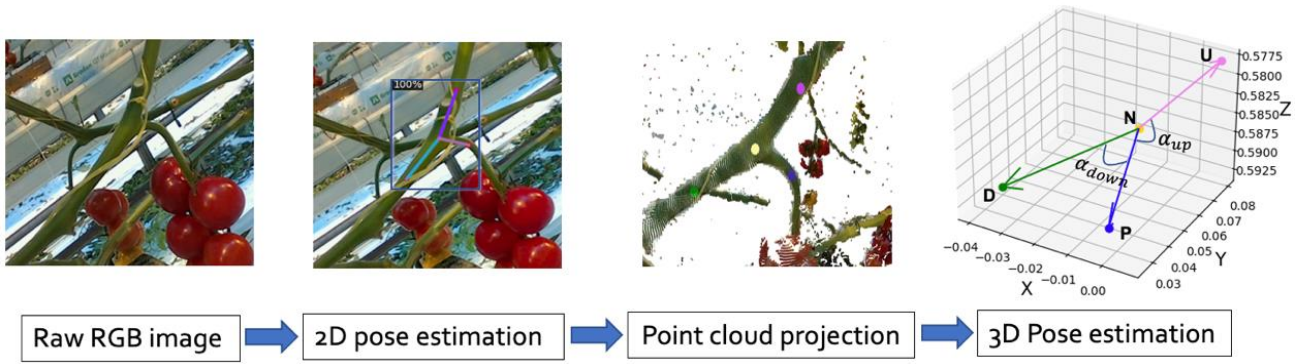

Figure 1. Three stages, keypoint detection, 3D point cloud projection, and 3D pose estimation.

## 2. Materials and Method

### 2.1 Materials

#### 2.1.1 Camera setup and dataset acquisition

The data were collected at a commercial tomato greenhouse located in Beek en Donk, Netherlands. This greenhouse is equipped with a fully automatic climate control system, including temperature, humidity, and lighting control. The tomato plants were grown in a high-wire pattern and the main stem was trained to grow vertically along the wire. The tomato variety used in this research was 'Tasty Tom', which has around 8 fruits per tomato truss, a peduncle diameter of around 7mm at maturity. Standard operational management was applied, including the de-leafing of the lower leaves surrounding the ripe tomato trusses.

A RealSense L515 LiDAR camera was used to capture both RGB images and aligned point clouds, both of which had a resolution of 1280x720. The camera was connected to an Alienware M15 R3 laptop with an Intel i7 10750H CPU and Nvidia GeForce RTX-2070s GPU, running on Windows 10 using the Pyrealsense2 package for image acquisition. Data was collected under ambient lighting, with the camera being hand-held between 50cm-80cm from a peduncle node, which was usually centered in the image. In addition to this node, a large number of peduncle nodes from other plants were present in the background of each image. Data collection occurred between October to December, in the 6th, 10th, and 11th weeks after planting.

The dataset included 648 RGB images, out of which 503 were used for model training and 145 for testing (Table 1). The training set included 614 annotated peduncle nodes, with 1-3 nodes per image, while the test set contained 145 annotated peduncle nodes, with a single peduncle node per image. Apart from the peduncle nodes, also leaf nodes (petiole node) were visible in the images. The detection algorithm had to ignore those and detect only the peduncle nodes.

The test set contained a subset of 95 images of 19 nodes from five different viewpoints to be able to analyze the detection performance for different poses of the peduncle in the image. These viewpoints were named V1 (canonical view), V2 (higher view), V3 (lower view), V4 (leftwards view), and V5 (rightwards view), all directed toward the peduncle node (see Figure 2). The positions of V2-V5 were roughly determined and had an angle of approximately $60° ± 15°$ from the straight perpendicular direction.

Table 1. The composition of the data set.

|  | **Total RGB** | **Total Nodes** | **Align point cloud** |
|---|---|---|---|
| **Training set** | 503 | 614 (1-3 per image) | No |
| **Test set** | 145 | 145 | Yes |

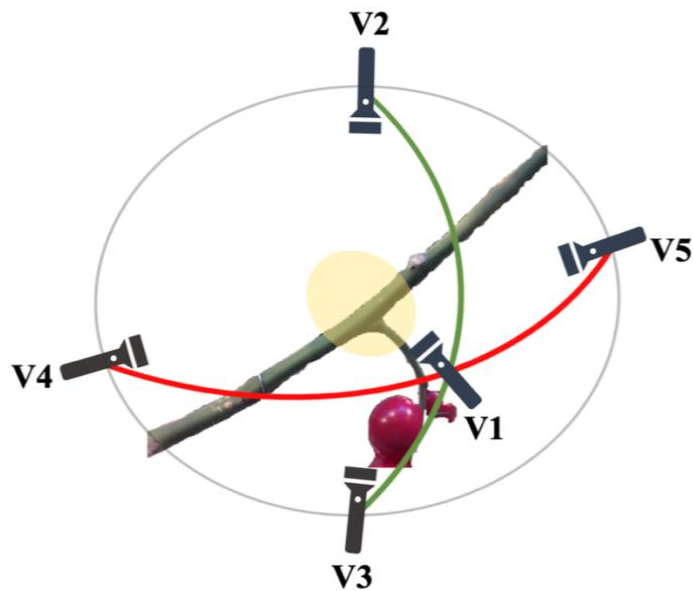

Figure 2. The distribution of the 5 viewpoints: V1 represents the canonical view, V2 represents the higher view, V3 represents the lower view, V4 represents the leftwards view, and V5 represents the rightwards view. All viewpoints are directed towards the peduncle node.

## 2.1.2 Ground-truth annotation in 2D

The RGB images were annotated using COCO annotator, an open-source web-based image annotation tool for object detection, segmentation, and keypoint detection (Brooks, 2019). As shown in Figure 3, four keypoints were labeled to express the pose of the peduncle node, where three points 'U', 'N', and 'D' were set at the upper, node, and lower positions of the main stem, respectively, and one point 'P' was set on the peduncle where it begins to bend. Only the visible keypoints were labeled. It is important to note that due to the lack of visible features, the positions of points 'U' and 'D' were determined based on Euclidean distance to 'N' at 80 pixels. The distance was determined arbitrarily and did not consider the distance from the camera lens to the object.

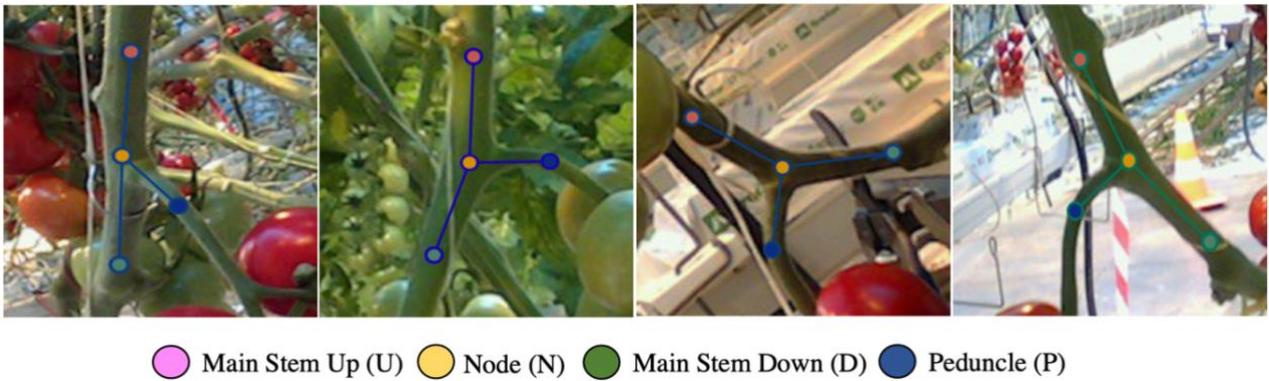

Figure 3. Samples of annotated images. Four keypoints were annotated to express the pose of the peduncle node. The 3 points 'U', 'N', and 'D' were set at the upper, node, and lower positions of the main stem, while 1 point 'P' was set on the peduncle.

## 2.1.3 Ground-truth measurement in 3D

To evaluate the pose-estimation method, we measured the ground-truth angles between peduncle and main stem manually. As shown in Figure 1, the upper angle between the peduncle and the main stem was defined as $\alpha_{up}$, which was calculated as $\alpha_{up} = \angle(\overrightarrow{NP}, \overrightarrow{NU})$, while the lower angle was defined as $\alpha_{down}$, calculated as $\alpha_{down} = \angle(\overrightarrow{NP}, \overrightarrow{ND})$. Three people measured the angles with protractors, and the average value was used. The standard deviations between the observers for all node measurements ranged between 1.73º-7.64º for $\alpha_{up}$, and between 0º-8.66º for $\alpha_{down}$.

## 2.2 Peduncle node pose estimation

### 2.2.1 Overview

As shown in Figure 1, the pose estimation system was designed with three stages: 1) 2D pose estimation, 2) point cloud projection, and (3) 3D pose estimation. The system takes the 2D color images and the aligned point cloud as input and estimates the 3D pose of the peduncle node as output. The first stage uses a deep keypoint detector on color images to localize the 4 pre-defined keypoints ('U', 'N', 'D', 'P'). In the second stage, the 2D keypoints are projected onto the point cloud to convert them into 3D Cartesian space. Finally, the 3D poses are estimated in the third stage. More details about the three stages are provided in the next subsections.

### 2.2.2 Pose estimation in 2D

#### *Keypoints R-CNN architecture*

This study employed the Keypoints R-CNN network (He et al., 2017), based on the Detectron2 implementation (Wu et al., 2019), for pose estimation on color images. Keypoint R-CNN takes a camera image as input and predicts the bounding box of each peduncle node, including the location of the four keypoints in the image. As shown in Figure 4, the network is comprised of two stages. In the first stage, a ResNet50-FPN (50-layer Residual Network (He et al., 2016) combined with a Feature Pyramid Network (Lin et al., 2017) was used as backbone to extract and fuse features from the input image. These feature maps were then processed by a region proposal network (RPN) (Ren et al., 2015) to generate a large set of regions of interest (ROI) that possibly contain target objects. In the second stage, the ROI proposals were used to crop the feature maps, which were subsequently processed by two independent fully connected (FC) networks: (1) an object-detection network to predict the bounding box of the object, along with a class labels and confidence score; and (2) a keypoint-detection network to predict the location of the keypoints of the detected object. The model originally had an additional branch for instance segmentation predicting the object mask, but it was removed as it was not used in this research.

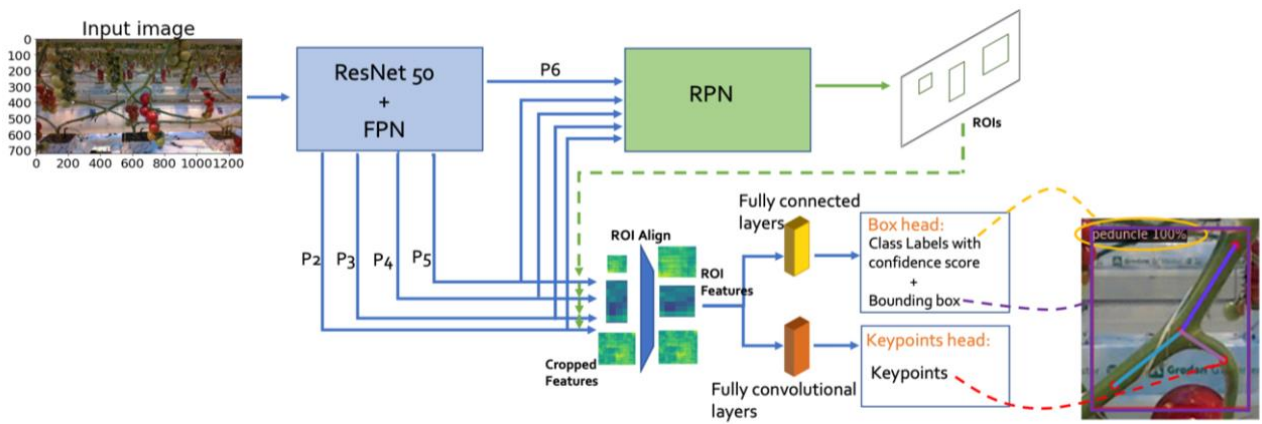

Figure 4. The structure of the Keypoints-RCNN algorithm. ResNet50-FPN is used to extract features, and RPN is utilized to generate ROI proposals by taking feature maps as input. The cropped ROI features are fed into two branches to predict class labels, bounding boxes, and keypoints.

#### *Training of the network*

The model was trained on a PC with an Nvidia-GPU RTX-3090 and Intel-CPU i9-10920X on Linux

Ubuntu 18.04, utilizing a training set of 503 images, as described in section 2.1.1. To speed up the process, transfer learning was applied by using a pre-trained network on the COCO dataset (Lin et al., 2014) and fine tuning on our training set. The training process consisted of 30,000 iterations with a batch size of 8, resulting in approximately 477 epochs. The Stochastic Gradient Descent (SGD) optimizer was used with a learning rate of 0.005 and a decay rate of 0.1 to optimize the network weights and biases to reduce the loss. During training, an anchor was considered positive if it had the highest intersection-over-union (IoU) or an IoU higher than 0.7, while anchors with IoU less than 0.3 were considered negative. To reduce the duplication of detections corresponding to the same ground truth, the Non-Maximum Suppression (NMS) was set to 0.7. During training, augmentation methods were employed to enhance the robustness of the model, including brightness with a random factor between 0.8-1.5, contrast with a random factor between 0.6-1.3, saturation with a random factor between 0.8-1.4, lighting through adding color jittering with a random degree sampled from normal distribution with a standard deviation of 0.7, and horizontal flipping with a probability of 50%. The default anchor sizes of 32 and 512 were removed as the peduncle nodes in our images were typically around 200 pixels in height and 140 pixels in width.

### *Peduncle-node detection*

The trained keypoint detector has the capability to localize the peduncle nodes and their associated keypoints in view. However, in practical applications, the harvesting robot can only harvest one tomato truss at a time. Hence, when multiple nodes were detected, the best candidate was determined based on the confidence value corresponding to the detection (Figure 5a).

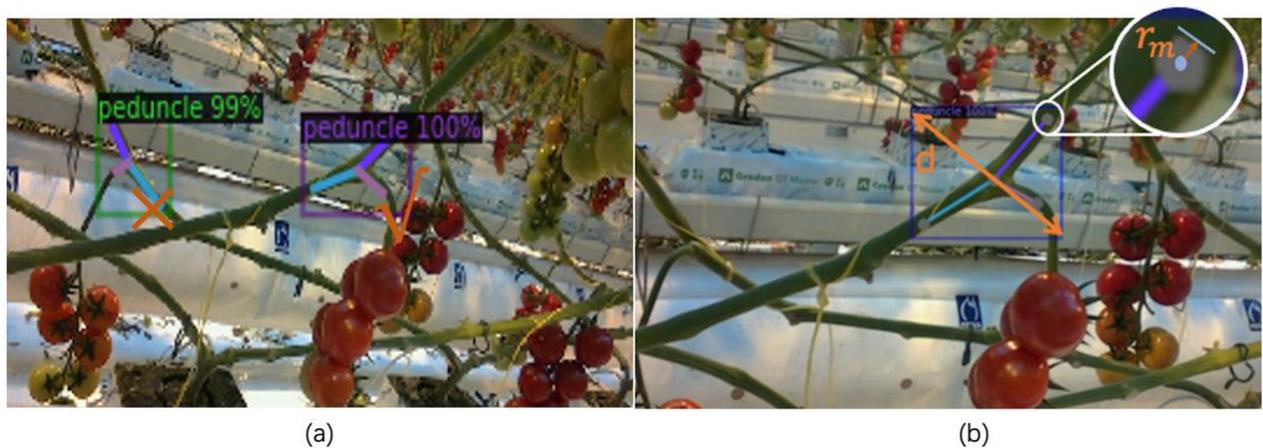

Figure 5. In figure (a), the detection with the highest confidence value was determined as the final target. In figure (b), the predicted keypoints of the selected target were extended to circular masks to include more pixels to be projected in the point cloud.

### 2.2.3 Pose estimation in 3D

*2D Keypoints Extension*

After a peduncle node was detected, its keypoints were projected into 3D by using the spatial information from the aligned point cloud. However, the collected point clouds are usually sparse, the projection of only the four keypoints might fail due to missing corresponding points in the point cloud. To deal with this, the 2D keypoints were extended with circular masks to include more pixels and all pixels inside the masks were projected to the point cloud (Figure 6a-b). To avoid segmenting extra pixels from the background, the radius $r_m$ of the circular masks was determined using Equation (1), where $d$ is the diagonal of the predicted bounding box and $w$ is a ratio set to 0.03 in our experiments. This ensured that the ratio of the circular mask was always determined relative to the size of the peduncle node.

$$r_m = w \cdot d \quad [pixel] \quad (1)$$

*Point cloud projection*

The color and depth images captured by the camera were aligned. Using the depth data, the 3D coordinates, $X_i, Y_i, Z_i$, were determined for all pixels $p_i$ that had valid depth data. Thus, the relation between the image coordinates, $p_i = \{x_i, y_i\}$, and the 3D coordinates was known. Figure 6a displays an example of a point cloud where the pixels within the keypoint masks in the color image were projected into the point cloud. However, it can be seen in Figure 6b-c that the raw point clusters were influenced by the outliers in the background. This caused the centroid of some clusters to deviate, see Figure 6c for an example. Outlier points were removed in the next step.

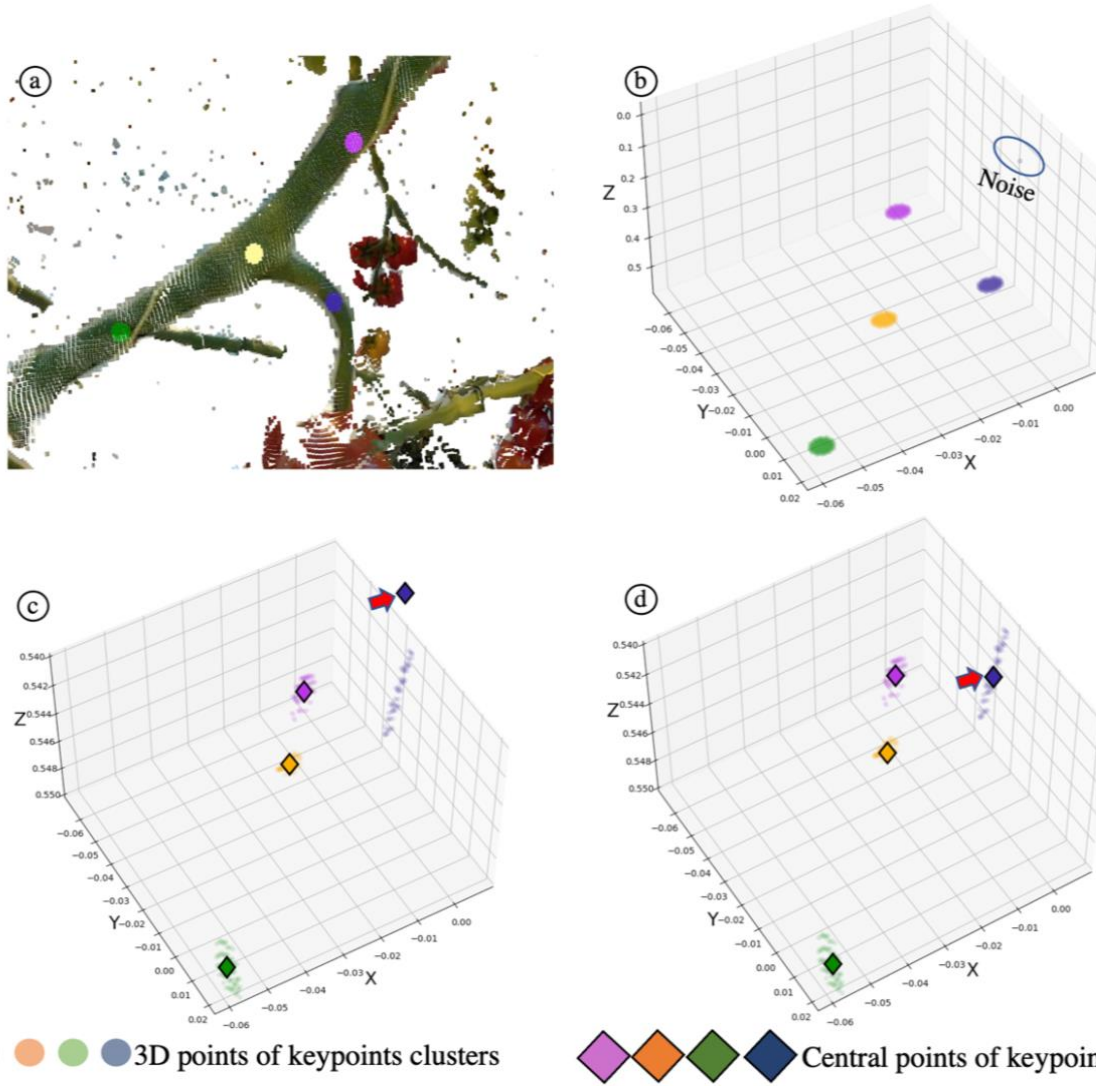

Figure 6. From 2D keypoints to 3D pose. a) show the masked keypoints in the color image projected into the 3D point cloud, with each keypoint marked in a different color. b) show the 3D points belonging to the keypoint clusters, containing outliers. c) show a zoom-in of b) with the centroids of each noisy clusters marked by diamonds. d) shows the centroids of the four clusters after removal of outliers.

### *Outliers Removal*

The outliers were determined based on the density of the surrounding points. A point was considered an outlier using Equation (2). For each point $i$, the number of neighbors within a 3D sphere of radius $r_f$ was counted. If there were fewer than the minimum required number of neighbors $n_{min}$, the point was handled as an outlier and removed. Consider $n_p$ as the total number of points and $d_{ij}$ as the distance from neighboring point $j$ to point $i$. The function $\rho$ equals 1 when $d_{ij} < r_f$, otherwise it equals 0. During the experiments, we used the radius $r_f = 0.005\ [m]$ and the threshold $n_{min} = 10$. These values were determined by a simple pilot experiment and by manually checking the result of noise removal.

$$Outlier_i = \begin{cases} 1, & \sum_{j=1}^{n_p} \rho\left(d_{ij} < r_f\right) < n_{min} \\ 0, & \text{otherwise} \end{cases} \quad (2)$$

## 2.3 Evaluation

A test set containing 145 RGB-D images was used to evaluate the performance of the algorithm from 3 perspectives: 1) object detection, 2) keypoint detection, and 3) pose estimation. In evaluations 1) and 2), only the RGB images were used, and in evaluation 3), the aligned point clouds resulting from the depth images were used. Most of the parameters for the algorithm in the testing phase were the same as in the training phase, except the maximum-detection-per-image parameter, which was modified to 1 corresponding to the detection with the highest confidence. In other words, the evaluation was done for the most prominent peduncle node in the image.

### 2.3.1 Evaluation of object detection

The objective of this experiment was to evaluate the algorithm's performance in detecting the peduncle nodes. A commonly used metric for object detection, the Average Precision (AP) and F1-socre, was applied to evaluate the performance. To judge if the detections were correct, a criterion of Intersection-over-Union (IoU) was employed to calculate the similarity between the predicted bounding box ($B_p$) and the ground-truth box ($B_{gt}$). It is by dividing the area of the intersection of the two bounding boxes by their union area:

$$IoU(B_P, B_{gt}) = \frac{area(B_p \cap B_{gt})}{area(B_p \cup B_{gt})} \quad [-] \quad (3)$$

The IoU ranges from 0 to 1, where a higher value represents a better match. Based on the calculated IoU, it is determined if a detection is a true positive (TP) or a false positive (FP), and if there are false negatives (FN). Specifically, if $IoU(B_P, B_{gt}) > \theta$, where $\theta$ is a predefined threshold, then the prediction is TP. A detection is FP if the $IoU(Bp, Bgt) \leq \theta$ or the annotated object is already detected by another prediction with a higher IoU value. If an annotated object is not detected, in other words, the IoU with all the predictions is lower than $\theta$, then it is an FN. Based on the number of TP, FP, and FN, the value of precision and recall under $\theta$ can be calculated. The value of precision represents what fraction of the detections were correct out of all the detections while recall reflects what fraction of the targets were correctly detected.

$$Precision^\theta = \frac{TP^\theta}{TP^\theta + FP^\theta} \quad [-] \quad (4)$$

$$Recall^\theta = \frac{TP^\theta}{TP^\theta + FN^\theta} \quad [-] \quad (5)$$

Based on the value of precision and recall, the AP and F1-score can be calculated. According to Equation (6), AP is determined as the area under the interpolated precision-recall curve, where $R_1, R_2, \ldots, R_n$ are the recall levels, and interpolated precision $P_{interp}(R)$ is the maximum precision found at any recall level $\geq R$. F1-score can be calculated using Equation (7), which computes the harmonic mean of the precision and recall giving a more intuitive evaluation for the performance. Both the evaluations were conducted using three $IOU$ thresholds $\theta$ at 0.5, 0.75, and 0.5:0.95.

$$AP@\theta = \sum_{i=1}^{n}(R_i^\theta - R_{i-1}^\theta)P_{interp}^\theta \quad [-] \tag{6}$$

$$F1-score@\theta = 2 \cdot \frac{Precision^\theta \cdot Recall^\theta}{Precision^\theta + Recall^\theta} \quad [-] \tag{7}$$

### 2.3.2 Evaluation of keypoint detection

This experiment aimed to evaluate the performance of the keypoint detector in detecting pre-defined keypoints of the peduncle nodes. Four keypoints were expected to be detected for each node. The detection performance was evaluated using the Percentage of Detected Joints (PDJ) metric, also known as Percentage of Correct Keypoints (PCKh) in some studies (Andriluka et al., 2014; X. Li et al., 2019; Russello et al., 2022). Equation (8) was used to calculate the PDJ. A predicted keypoint was considered correctly detected if the Euclidean distance ($d_i$) between it and the ground truth was smaller than a certain fraction $f$ of the diagonal $d$ of the predicted bounding box. Accounting for the object diagonal allowed the method to handle nodes of arbitrary sizes. The term $n$ referred to the number of keypoints on a node. And $\sigma(x) = 1$ if $x \leq 0$, otherwise =0.

$$PDJ@f = \frac{1}{n}\sum_{i=1}^{n}\sigma(d_i - f*d) \quad [\%] \tag{8}$$

In this analysis, the PDJ values were calculated for three fractions $f$: 0.05, 0.1, and 0.2. The smaller the value of $f$, the closer the predicted points need to be to the ground truth keypoints to be considered correct. The analysis was conducted from two perspectives: (1) detection-wise and (2) keypoint-wise. The detection-wise evaluation calculated the number of detections that achieved different PDJ values. The PDJ of a detection was 1 if all 4 keypoints were correctly detected, and 0 if none of the keypoints were detected correctly. The keypoint-wise evaluation calculated success rate per keypoint type, with a higher ratio indicating a higher successful detection rate.

### 2.3.3 Evaluation of pose estimation

Using the 3D keypoints generated in section 2.2.3., the 3D pose of the peduncle node was estimated. The 3D pose was defined to include the node position, orientation, and relative angles. However, measuring the complete ground truth was challenging, so only the relative angles $\alpha_{up}$ and $\alpha_{down}$ were evaluated. The angels were calculated using Equation (9):

$$\alpha = arccos \frac{\vec{A} \cdot \vec{B}}{|A| \cdot |B|} \tag{9}$$

for any two given vectors $\vec{A}$ and $\vec{B}$ in 3D space, their included angle $\alpha$ equals their dot product divided by their magnitudes. For 4 keypoints 'U', 'N', 'D', and 'P', the $\alpha_{up}$ was calculated using vectors $\vec{NU}$ and $\vec{NP}$, while $\alpha_{down}$ was calculated by $\vec{ND}$ and $\vec{NP}$.

We considered two primary sources of error in the estimated angles: (1) pose estimation in 2D, resulting from incorrect localization of keypoints, (2) pose estimation in 3D, including 3D keypoints projection, noise filtering, and pose estimation. We did not consider errors resulting from limitations in the sensor's capability, such as insufficient depth measurements leading to bad quality of generated point cloud, as these would not reflect the performance of the algorithm. After removing the data where 3D information was not adequately captured, 134 test samples remained for analysis.

To gain a deeper understanding of the source of errors, we evaluated the accuracy of three types of predicted angles: (1) $\hat{\alpha}^{\mathrm{anno}}$ is the predicted angle based on the annotated keypoints, so with perfect keypoint detection, which reflect errors caused in 3D, (2) $\hat{\alpha}^{\mathrm{kp}}$ is the predicted angle based on the predicted keypoints, which includes errors in keypoint detection and 3D calculation, and (3) $\hat{\alpha}^{\mathrm{4kp}}$ represents the predicted angles only when all 4 keypoints were correctly detected, which reflects the optimal performance that our method can achieve. To compare to, $\alpha$ represents the ground-truth angle manually measured in the greenhouse.

Two evaluation metrics, Mean Absolute Error (MAE) and Mean Relative Error (MRE), were used to evaluate the accuracy of the estimated angles. The MAE is calculated as the mean of the absolute errors between the predicted angle, $\hat{\alpha}_t$, and the ground-truth angle, $\alpha_t$, for peduncle $t$, as shown in Equation (10). The MRE was calculated as the mean of the relative errors obtained by dividing the absolute errors with the corresponding ground-truth angle, as shown in Equation (11). Here, $T$ denotes the total number of detections.

$$\mathrm{MAE} = \frac{1}{T} \sum_{i=0}^{t} |\hat{\alpha}_t - \alpha_t| \quad [°] \tag{10}$$

$$MRE = \frac{1}{T} \sum_{i=0}^{t} \frac{|\hat{\alpha}_t - \alpha_t|}{\alpha_t} \quad [\%] \tag{11}$$

## 3. Results

The results of the evaluation in terms of object detection, keypoint detection, and 3D angle estimation were displayed in sections 3.1, section 3.2, and section 3.3 respectively.

## 3.1. Node detection

Table 2 presents the overall AP and F1-score for different IoU thresholds (0.5, 0.75, and

0.5:0.95). The method achieved outstanding results for both metrics on $IoU_{0.5}$, with AP@0.5=0.96 and F1-score@0.5=0.98, indicating the model can successfully detects most of the peduncles under a relatively tolerant intersection-over-union criterion. However, both metrics experienced a significant decrease for a stricter threshold ($IoU_{0.75}$), suggesting that the bounding-box predictions of the complete peduncle were not very accurate with a mismatch to the ground-truth bounding box. It should be noted that the peduncle node, as we defined it, is poorly represented by a bounding box, so it is not unexpected that the accuracy of the bounding-box prediction is not so high. We do not use the bounding-box predictions for further operations, but instead use the keypoint detections discussed in the next subsection.

*Table 2. The overall results of AP and F1-score.*

| Metrics | $IoU_{0.5}$ | $IoU_{0.75}$ | $IoU_{0.5:0.95}$ |
|---|---|---|---|
| AP | 0.96 | 0.37 | 0.45 |
| F1-score | 0.98 | 0.53 | 0.54 |

According to Table 3, the algorithm accurately detected nodes from all 5 viewpoints, with an AP@0.5 greater than 0.91 and an F1-score@0.5 greater than 0.95 for all viewpoints, reaching 1.0 for three of the five viewpoints. This suggests the algorithm can deal with the variations caused by different viewpoints. Specifically, Viewpoints 1, 2, and 3 achieved perfect node detection, while Viewpoints 4 and 5 had lower accuracy. This is likely because the lateral views can introduce more shape variations in the peduncle node, making it more difficult for the algorithm to accurately detect. When looking at $IoU_{0.75}$ and $IoU_{0.5:0.95}$ thresholds for both the metrics, we see larger differences. Viewpoint 2 exhibited the highest values across all metrics, indicating that the top view is the optimal viewpoint for tasks such as harvesting. Viewpoint 3 exhibited a significantly lower value. One possible reason could be the impact of lighting on the images, as this viewpoint is more susceptible to lighting variations, which affects the algorithm's ability to detect the bounding-box of the peduncle nodes accurately.

*Table 3. The AP and F1-score obtained from different viewpoints.*

| Viewpoint | AP | | | F1-score | | |
|---|---|---|---|---|---|---|
| | $IoU_{0.5}$ | $IoU_{0.75}$ | $IoU_{0.5:0.95}$ | $IoU_{0.5}$ | $IoU_{0.75}$ | $IoU_{0.5:0.95}$ |
| View1 | 1.0 | 0.33 | 0.48 | 1.00 | 0.47 | 0.54 |
| View2 | 1.0 | 0.86 | 0.67 | 1.00 | 0.89 | 0.71 |
| View3 | 1.0 | 0.08 | 0.30 | 1.00 | 0.16 | 0.37 |
| View4 | 0.94 | 0.47 | 0.46 | 0.95 | 0.58 | 0.52 |
| View5 | 0.91 | 0.17 | 0.35 | 0.95 | 0.21 | 0.38 |

### 3.2. Keypoint detection

The evaluation of detected keypoints was performed on three fractions ($f$) of 0.05, 0.1, and 0.2. Table 4 displays the ratio of the detections that achieved different PDJ values, a higher PDJ value indicates more keypoints were correctly detected. The PDJ@0.2 and PDJ@0.1 display a similar distribution, with the most of detections achieving a PDJ of 100% (85.52% for PDJ@0.2 and 46.90% for PDJ@0.1), while the smallest proportion of detections achieved a PDJ smaller than 50% (2.07% for PDJ@0.2 and 6.12% for PDJ@0.1). In contrast, for PDJ@0.05, only 13.10% of detections achieved PDJ=100%, while a large number of detections (31.03%) achieved a PDJ smaller than 50%. The significant shift in the ratio when reducing $f$ from 0.2 to 0.05 is that the smaller fraction $f$ requires the detected keypoints be closer to the ground truth to be considered correct, which resulted in most of the detected keypoints considered correct under $f$=0.2 being considered as failed under a more stringent criterion ($f = 0.05$).

The results of PDJ@0.2 and PDJ@0.1 indicate the feasibility of the model, which essentially located the keypoints on the right locations. But some detections are not accurate enough in specific keypoints or scenarios to fulfill a more stringent criterion. To investigate the sources of error, further analysis was carried out by calculating the PDJ for each individual keypoint.

*Table 4. The number of images that achieved different PDJ values are counted at three fractions, 0.05, 0.1, and 0.02 independently.*

| PDJs | <50% | 50% (2/4) | 75% (3/4) | 100% (4/4) |
|---|---|---|---|---|
| PDJ@0.05 | 31.03%(45) | 37.35%(44) | 25.52%(37) | 13.10%(19) |
| PDJ@0.1 | 6.21%(9) | 23.45%(34) | 23.45%(34) | 46.90%(68) |
| PDJ@0.2 | 2.07%(3) | 2.07%(3) | 10.34%(15) | 85.52%(124) |

As shown in Table 5, the 4 keypoints 'U', 'N', 'D' and 'P' achieved PDJ@0.2 at 91.03%, 97.93%, 91.72%, and 96.55% respectively, suggesting the model was capable of accurately localizing the keypoints and achieving a success rate above 90% for all the keypoints. The corresponding PDJ values decreased at varying levels by 60%, 8.96%, 57.93%, and 35.17% when reducing $f$ from 0.2 to 0.05. The slight reduction of keypoint 'N' indicates it could be consistently localized close to the ground truth, and hence, its results were not largely affected even under a stricter criterion. In contrast, the significant decrease of keypoints 'U' and 'D' at PDJ@0.05 suggests these two keypoints usually deviated largely from the ground truth, and thus were no longer considered correct when $f$ was decreased to 0.05. The keypoint 'N' achieved the highest PDJ in all three fractions, implying the most accurate detection, while the keypoint 'U' and 'D' showed significantly lower success rates, and their gap with the keypoint 'N' became even larger when decreasing $f$.

As described in section 2.1.2, keypoints 'N' and 'P' were determined based on unique visual features ('N' linking the peduncle and the main stem; 'P' normally bending), while keypoints 'U' and 'D' were annotated based on the Euclidean distance. Based on the results, we state that the visual features are more effective for the model to learn and recognize compared to the distance features.

Table 5. The PDJ of different keypoints for the entire testset with 145 images under 3 fractions of 0.05, 0.1, and 0.2.

| PDJs | U | N | D | P |
|---|---|---|---|---|
| PDJ@0.05 | 31.03% | 88.97% | 33.79% | 61.38% |
| PDJ@0.1 | 57.93% | 97.93% | 60.69% | 92.41% |
| PDJ@0.2 | 91.03% | 97.93% | 91.72% | 96.55% |

Figure 7 displays some samples in which the localizations of the keypoints 'U' or 'D' were considered incorrect due to large deviations from the ground truth. However, all these localizations were found to be correctly assigned to the corresponding plant parts. Based on this, we state that the results can be further improved if the system is tasked with accurately locating the different landmarks without necessarily requiring proximity to the ground truth.

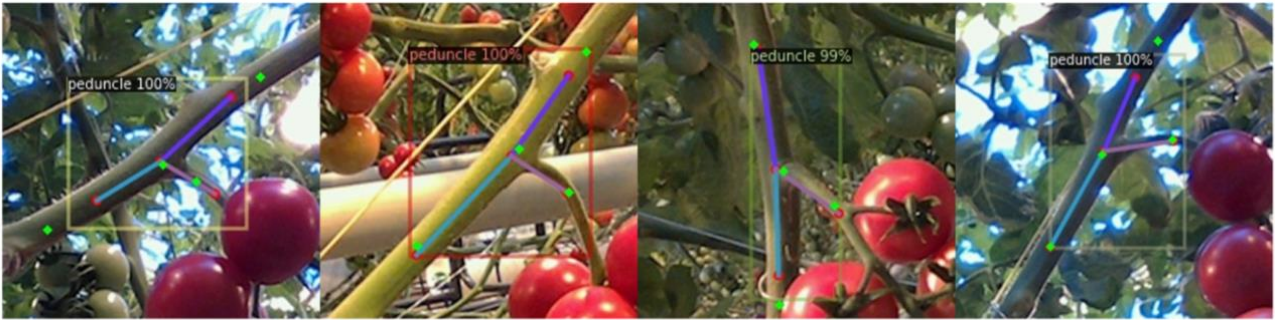

Figure 7. Some detections failed to localize the keypoints 'U' and 'D' due to their larger deviations from the ground truth. The green squares represent the ground truth keypoints.

Table 6 presents keypoints detection results from different viewpoints, with an average PDJ@0.2 of over 0.89 for all views and two views reaching above 0.97. View1 and View2 outperformed the other views, making them ideal for keypoints detection in practical applications. Keypoints 'N' and 'P' consistently showed significantly higher PDJ values compared to 'U' and 'D' across all views. Notably, View5 displays lower PDJ for 'P'. This is because in our greenhouse environment, the peduncles tend to grow towards the same direction, which is roughly View5, leading to significant shape variations and challenging keypoints detection. The reduced detection accuracy of 'P' can largely impact angle estimation, as its precise 3D position is crucial for estimating both angles, $a_{up}$ and $a_{down}$.

Table 6. The PDJ obtained by the algorithm at different viewpoints under fraction equals 0.2.

| Viewpoints | U | N | D | P | Mean |
|---|---|---|---|---|---|
| View1 | 1.00 | 1.00 | 0.95 | 1.00 | 0.99 |
| View2 | 0.95 | 1.00 | 0.95 | 1.00 | 0.97 |
| View3 | 0.74 | 1.00 | 0.95 | 1.00 | 0.92 |
| View4 | 0.89 | 0.95 | 0.79 | 0.95 | 0.89 |
| View5 | 0.95 | 0.95 | 0.89 | 0.89 | 0.92 |
| Mean | 0.91 | 0.98 | 0.91 | 0.97 | |

## 3.3. 3D Pose estimation

Based on the results presented in Figure 8, $\hat{a}^{anno}$ achieved a MAE of 8.25° and 7.21° for $\alpha_{up}$ and $\alpha_{down}$, respectively, with a MRE of 12.12% and 5.99%. These errors were mainly caused by pose estimation in 3D, indicating that even when all keypoints are accurately localized, errors can still occur. These errors serve as a minimum level of error that can be achieved by the algorithm. $\hat{a}^{kp}$, which includes the errors caused by pose estimation in both 2D and 3D, achieved larger errors with an MAE of 11.38° and 9.93° for $\alpha_{up}$ and $\alpha_{down}$, respectively, and an MRE of 16.68% and 8.29%. The larger errors in $\hat{a}^{kp}$ suggest that a significant proportion of the errors were caused by pose estimation in 2D. So, in the evaluation of $\hat{a}^{4kp}$, the errors were reduced to similar levels to $\hat{a}^{anno}$. The slightly larger error of $\hat{a}^{4kp}$ than $\hat{a}^{anno}$ is because the predicted keypoints and the annotated keypoints usually deviated, and these slight deviations could cause errors.

Compared to $\alpha_{down}$, $\alpha_{up}$ showed a slightly larger MAE but a significantly larger MRE. This is because the actual angle of $\alpha_{up}$ is generally much smaller, and as a result, the MRE can be amplified when the MAE is similar to that $\alpha_{down}$.

It is important to note that not all 134 detections were considered in the evaluation of $\hat{a}^{kp}$, as there were cases where $\alpha_{up}$ or $\alpha_{down}$ were incomputable. For example, keypoints localized on the background due to incorrect keypoints detection usually lack 3D information, which can make the angles incomputable. In this evaluation, only one case was excluded for this reason, so a total of 133/134 cases were included. For $\hat{a}^{4kp}$, 115 detections were used.

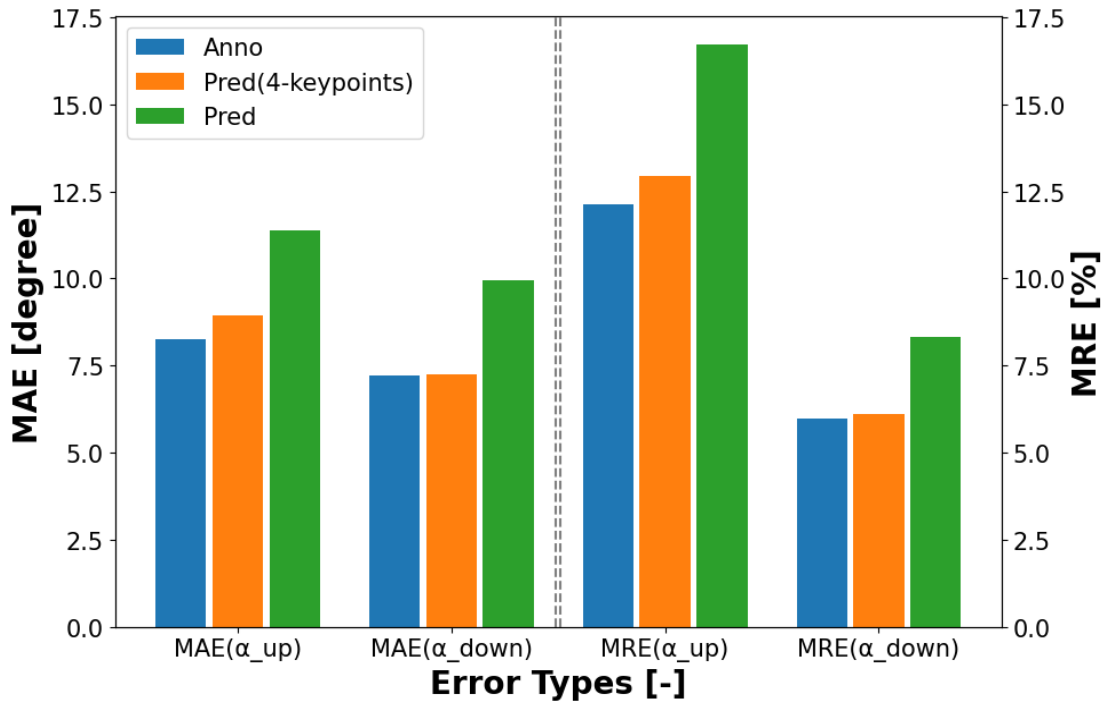

Figure 8. Three kinds of angles were evaluated on MAE and MRE. The 'Anno' represents the errors contained in $\hat{a}^{anno}$, the 'Pred' represents the errors contained in $\hat{a}^{kp}$, while 'Pred(4-keypoints)' represents the errors contained in $\hat{a}^{4kp}$ only when 4 keypoints are correctly detected

under $f$=0.2.

Table 7 displays the results of angle estimation acquired from different viewpoints. View1 showed the lowest errors, followed by View2 and View3, which had a vertical deviation from View1, lower than View4. However, View5 exhibited significantly higher errors, mainly due to the low detection accuracy of 'P' at this view (refer to Section 3.2), suggesting that this view should be avoided in real applications.

*Table 7. Errors achieved by $\hat{a}^{kp}$ in angle estimation at different viewpoints.*

| Viewpoint | MAE (up) | MAE (down) | MRE (up) | MRE (down) |
|---|---|---|---|---|
| View1 | 9.13° | 6.17° | 7.49% | 5.05% |
| View2 | 11.27° | 9.36° | 10.07% | 7.60% |
| View3 | 9.16° | 6.90° | 7.55% | 5.62% |
| View4 | 12.09° | 11.09° | 9.30% | 8.50% |
| View5 | 21.30° | 23.02° | 17.51% | 18.75% |

## 4. Discussion

### 4.1 General evaluation of proposed method

Our method demonstrated a high success rate in detecting the peduncle nodes. Although the nodes were detected well, the exact bounding-box coordinates often did not accurately match the ground truth. This is mainly because keypoints 'U' and 'D' were annotated based on distance and not based on a clear image feature, which the model struggled to effectively learn. This was further evident in the evaluation of the keypoint detection. While the method displayed high accuracy across all types of keypoints, 'U' and 'D' showed significantly lower PDJ compared to 'N' and 'P'. Since our method directly performs 3D pose estimation based on the detected keypoints, the accuracy of the 3D pose estimation relied on the accuracy of the 2D pose detection.

Our model demonstrated capability in handling variations caused by different viewpoints with respect to the peduncle node. The node detection results showed an AP ranging between 0.94-1.00, while the F1 score ranged between 0.95-1.00, indicating high detection performance for all viewpoints, with View 1-3 showing the best performance. Similarly, the results of the keypoint detection showed a high consistency over viewpoints, with PDJ between 0.89 and 0.99, with View 1 and 2 resulting in the best performance and View 4 showing the lowest performance. The accuracy of the pose estimation showed more variation between viewpoints, with errors ranging between 5° and 12° for Views 1-4 and errors between 18° and 21° for View 5. Overall, View 1 and 2 showed to provide the best view on the node. While the node and keypoint detection for View 5 were good in general, it did show a lower PDJ for the node point 'P'. As the 3D location of this point is crucial to estimate the two pose angles, this explains the lower pose-estimation accuracy for this view. View 5 is the rightwards view, which relates to the view where the tomato truss is closer to the camera than the node point 'P', possibly partially blocking a good view on the node.

## 4.2 Future improvements

To evaluate the algorithm's performance in a realistic scenario, all test data were collected from a natural greenhouse. However, the complex environment posed challenges for object annotation and ground-truth measurement, as it was impossible to cover all the nodes in the background. To find a solution, only the main node in each image was considered, matching the prediction that gained the maximum confidence value. Focusing only on the most confident prediction ignored other predictions based by the method. However, for the use of robotic harvesting, that is justified, as the robot will harvest the trusses one by one anyway. Occasionally, our method predicted a node in the background, which was evaluated as a FP. A similar problem was seen in the experiment of Rong et al. (2022), where the models were trained to detect only the fruits and peduncles of tomato plants in the foreground, but targets in the background were still likely to be detected because both had quite similar traits. In future work, depth information can be integrated as input for the pose-estimation method to ignore the nodes in the background. This would force the model to only focus on objects in a certain depth range.

As in normal in commercial production, the leaves surrounding the peduncle node were removed before the fruit ripened. Consequently, situations where the peduncle node was occluded by the leaves were not considered. The deep-keypoint-detection method can inherently handle a certain level of occlusions and estimate the object's pose based on a subset of the visible keypoints. However, if the nodes are significantly occluded, we believe the current method may encounter some limitations. In such cases, utilizing the information from multi-views is a popular solution, it makes information that is not perceivable from a single view become available, and reduces perception uncertainty (Hemming et al., 2014; Rapado-Rincón et al., 2023; Shi et al., 2019; van Henten et al., 2003). To increase efficiency, next-best-viewpoint (NBV) methods can be used that propose camera poses to maximize the information gain (Burusa et al., 2022; Mendoza et al., 2020; Zeng et al., 2020). As the results showed, some viewpoints are more favorable for a good peduncle-node detection than others. In future work, we propose to integrate methods for NBV planning with our keypoint-based pose estimator to propose camera poses that maximize the view on the peduncle.

Due to the lack of visual features, the keypoints 'U' and 'D' were determined based on pixel distance, which was expected to be learned by the model. However, as shown in Table 5, these two keypoints presented significantly lower PDJ values compared to keypoints 'N' and 'P'. Figure 7 also suggests that keypoints 'U' and 'D' were likely considered incorrect due to their large deviations from the ground truth, even though they were localized on the correct plant part and could be used to calculate the angles. So, we consider the traditional PDJ metric may not the most suitable way to evaluate these two keypoints. On the other hand, keypoints are possible not the best possible representation for these points. Instead, the DNN could be adapted to predict a vector up the main stem and a vector down the main stem instead of two keypoints.

In section 2.2.3, circular masks with a radius of $r_m$ (Equation 1) were used to expand the keypoints to include more pixels. The value of $r_m$ was self-adjusted based on the diagonal ($d$) of the predicted bounding box and a constant weighting factor ($w$) to avoid segmenting the pixels in the background. This approach was found to significantly reduce errors in 3D pose estimation compared to using a single pixel. However, beyond a certain range (0-0.05), increasing the value of $w$ caused a negative effect, as pixels in the background were added. The sensitivity of the model to $w$ will reduce its robustness. We consider this problem can be solved by employing a better strategy for expanding keypoints including 3D distance information, or by growing only inside the mask of the node.

In Figure 8, the errors of the estimated angles did not reduce to very low even when all keypoints were accurately located ($\hat{a}^{anno}$ and $\hat{a}^{4kp}$), suggesting potential errors in the ground truth measurements. The measurement of $a$ was subjective as three measurers manually located keypoints and measured angles with a protractor. The large standard deviations in $\alpha_{up}$ and $\alpha_{down}$ (section 2.1.3) also highlight significant deviations in measurements. To improve accuracy and reduce time consumption, we recommend using $\hat{a}^{anno}$ directly as the ground truth in future experiments.

The analyses of the influence of the viewpoint with respect to the node suggests that some viewpoints are better than others. Particular view 1 (canonical view) and view 2 (the higher view) resulted in higher detection and pose-estimation performance. In future work, active-vision methods should be explored to actively guide the robot to the right viewpoint to make the most accurate detection.

### 4.3 Comparison with relevant studies

Boogaard et al. (2020) utilized YOLO v3 (Redmon & Farhadi, 2018) to detect cucumber internodes using multi-view imaging. Our study achieved comparable results if it comes down to the node detection, although a direct comparison is not possible due to differences in experimental conditions, tasks, and datasets. It is worth noting that their experiments were conducted in a controlled environment, while ours occurred in a commercial greenhouse, posing more challenges for node detection. They investigated the benefits of multiple viewpoints in object detection, which is what we can explore for keypoints detection method in future work. While their work only detected the nodes, our method also estimated the pose of the peduncle node.

Our keypoint detection results are consistent with Zhang et al. (2022), who used a DL-based keypoint detector for tomato bunches. Both studies found that keypoints 'U' and 'D' had significantly lower PDJ values compared to 'N' and 'P', supporting our assumption that distance is not an effective feature for model learning. While our study used a stricter evaluation criterion based on the weighted size of the node, they used the weighted size of the entire tomato bunch, allowing for greater deviation in their predicted keypoints to still be considered correct. Nevertheless, our method demonstrated higher PDJ values for 'N' and 'P', confirming its effectiveness in keypoints detection.

Our approach for 3D pose estimation was compared with Luo et al (2022), who used Mask R-CNN for target detection in RGB images and estimating peduncle orientations by mapping the detections to the aligned point clouds. Direct comparison of the results is not possible due to differences in evaluation metrics and targets. Their method focused solely on peduncle orientation, ignoring relative angles. Furthermore, our approach estimated the pose by processing only the keypoint clusters in point cloud, making it more efficient and simpler than their method, which processed the entire point cloud. However, their method, by using more complete point cloud can be less sensitive to 2D object detection accuracy and point cloud quality.

None of these three studies investigated the impact of different viewing angles on the algorithm performance, which provided valuable insights for selecting viewpoints in practical applications.

Our method demonstrated superior performance compared to similar studies, making it suitable for robotic harvesting tasks, with potential for further improvements to enhance accuracy. However, defining precise accuracy requirements for robotic harvesting poses challenges due to limited research and variations in crops, harvesting mechanisms, and other factors. For example, the design of grippers and cutters in a robotic harvester can tolerate some level of error, reducing the accuracy requirements for the vision system significantly (Zhang et al., 2022).

## 5 Conclusions

In this study, an innovative method was proposed to estimate the 3D pose of tomato peduncle nodes from RGB-D images. The method provided complete 3D pose information, including the position, orientation, and relative angles to the main stem, which are needed for robotic operations. Meanwhile, this method is expected to benefit other agricultural operations using robotics, such as phenotyping and monitoring, with the output of comprehensive information. After being analyzed from the three aspects of (1) object detection, (2) keypoint detection, and (3) pose estimation, the feasibility of our algorithm was shown. For each evaluation, the impact of viewpoint angles was investigated, indicating the model can handle a level of variations caused by view change. The evaluation resulted in the following conclusions:

(1) For object detection, our algorithm achieved outstanding results in both AP@0.5 and F1-score@0.5, suggesting its good capability in node detection. However, there was a significant drop in both metrics for the IoU threshold of 0.75, indicating that most of the bounding-box predictions do not accurately match the ground truth.
(2) Keypoint detection resulted in above 90% of PDJ@0.2 for all keypoints, demonstrating the model's capability to accurately localize the keypoints. However, there is a significant drop in PDJ@0.05 for 'U' and 'D' keypoints, indicating these two keypoints are localized with less accuracy. Nevertheless, as discussed earlier, some predicted keypoints considered incorrect can still be localized on the correct plant parts.
(3) For pose estimation, the errors of $\hat{a}^{4kp}$ displayed a significant decrease compared to $\hat{a}^{kp}$, indicating a large proportion of the errors in 3D pose estimation were due to incorrect keypoint detection in 2D. However, even if all the keypoints were accurately detected, there was still a level of error, suggesting the measurement of $a$ might not be very accurate. Therefore, in future experiments, the $\hat{a}^{anno}$ could be directly used as the ground-truth, which can save significant time in actual angle measurement.
(4) The methods achieved high AP@0.5 (>0.91) in object detection and PDJ@0.2 (>0.89) in keypoint detection for all viewpoints, indicating robustness to variations caused by view changes. View1 (canonical view) and view2 (higher view) yielded the best results and are recommended for practical applications.

While the proposed method was tested for the 3D pose estimation of tomato peduncles, the method can be applied to other similar crops, such as cucumber, zucchini, and bell pepper and for other plant parts, such as the pedicel (e.g., for deleafing) and fruit clusters (e.g., for monitoring tasks). Besides robotic applications, the method can also provide valuable information for plant monitoring and plant phenotyping.

## CRediT author statement

**Jianchao Ci**: Conceptualization, Methodology, Software, Investigation, Data Processing, Writing - Original draft; **Xin Wang:** Conceptualization, Data Processing, Writing - Review & Editing, Supervision; **David Rapado-Rincón:** Conceptualization, Software, Writing - Review & Editing; **Akshay K. Burusa:** Conceptualization, Writing - Review & Editing; **Gert Kootstra:** Conceptualization, Writing - Review & Editing, Supervision, Funding acquisition.

## Funding

This research is funded by the Netherlands Organization for Scientific Research (NWO) project Cognitive Robots for Flexible Agro-Food Technology (FlexCRAFT), grant P17-01.

## Declaration of competing interest

We declare that there are no financial and personal relationships that have inappropriately influenced the work reported in this paper.

## Acknowledgements

We thank the members of the FlexCRAFT project for engaging in fruitful discussions and providing valuable feedback to this work. Special thanks to Vereijken Kwekerijen for welcoming us to their greenhouse.